%% file: gsplat.tex
\documentclass[twoside,11pt]{article}

\usepackage[nohyperref]{jmlr2e}

\usepackage{blindtext}
\usepackage[svgnames, table]{xcolor}
\usepackage{amsmath}        
\usepackage{amssymb}        
\usepackage[symbol]{footmisc}       
\usepackage{listings}       
\usepackage{caption}        
\usepackage{tikz}           
\usepackage[noend]{algpseudocode}
\usepackage{booktabs}
\usepackage[colorlinks=true,allcolors=black]{hyperref}

\newcommand{\boldparagraph}[1]{\noindent{\bf #1} }

\definecolor{codegreen}{rgb}{0,0.6,0}
\definecolor{codegray}{rgb}{0.5,0.5,0.5}
\definecolor{codepurple}{rgb}{0.58,0,0.82}
\definecolor{backcolour}{rgb}{0.95,0.95,0.92}

\lstdefinestyle{codeblock}{
  backgroundcolor=\color{backcolour}, commentstyle=\color{codegreen},
  keywordstyle=\color{magenta},
  numberstyle=\tiny\color{codegray},
  stringstyle=\color{codepurple},
  basicstyle=\ttfamily\footnotesize,
  breakatwhitespace=false,         
  breaklines=true,                 
  captionpos=b,                    
  keepspaces=true,                 
  numbers=left,                    
  numbersep=5pt,                  
  showspaces=false,                
  showstringspaces=false,
  showtabs=false,                  
  tabsize=2,
}

\usepackage{lastpage}

\ShortHeadings{\texttt{gsplat}: An Open-Source Library for Gaussian Splatting}{\texttt{gsplat}}
\firstpageno{1}

\begin{document}

\title{\texttt{gsplat}: An Open-Source Library for Gaussian Splatting}


\author{%
  \name Vickie Ye$^{1,\dagger}$ \email vye@berkeley.edu\\
  \vspace{-0.6cm}
  \AND
  Ruilong Li$^{1,\dagger}$ \email ruilongli@berkeley.edu\\
  \vspace{-0.6cm}
  \AND
  Justin Kerr$^{1,*}$ \email justin\_kerr@berkeley.edu\\
  \vspace{-0.6cm}
  \AND
  Matias Turkulainen$^{2,*}$ \email matias.turkulainen@aalto.fi \\
  \vspace{-0.6cm}
  \AND
  Brent Yi$^{1,*}$ \email brentyi@berkeley.edu\\
  \vspace{-0.6cm}
  \AND
  Zhuoyang Pan$^{3,*}$ \email panzhy@shanghaitech.edu.cn\\
  \vspace{-0.6cm}
  \AND
  Otto Seiskari$^{4,*}$  \email otto.seiskari@spectacularai.com\\
  \vspace{-0.6cm}
  \AND
  Jianbo Ye$^{5,*}$ \email jianboye.ai@gmail.com\\
  \vspace{-0.6cm}
  \AND
  Jeffrey Hu$^{*}$ \email hujh14@gmail.com\\
  \vspace{-0.6cm}
  \AND
  Matthew Tancik$^{6,\dagger\dagger}$ \email matt@lumalabs.ai\\
  \vspace{-0.6cm}
  \AND
  Angjoo Kanazawa$^{1,\dagger\dagger}$ \email  kanazawa@eecs.berkeley.edu\\
  \vspace{-0.6cm}
  \AND
  {\addr $^1$ UC Berkeley}
  {\addr $^2$ Aalto University}
  {\addr $^3$ ShanghaiTech University}
  {\addr $^4$ SpectacularAI}
  {\addr $^5$ Amazon}
  {\addr $^6$ Luma AI}\\
  {\addr $^\dagger$Project Lead, $^*$Core Developer, $^{\dagger\dagger}$Project Mentor}\\
}

\editor{} 

\maketitle

\begin{abstract}
\texttt{gsplat} is an open-source library designed for training and developing Gaussian Splatting methods. It features a front-end with Python bindings compatible with the PyTorch library and a back-end with highly optimized CUDA kernels. \texttt{gsplat} offers numerous features that enhance the optimization of Gaussian Splatting models, which include optimization improvements for speed, memory, and convergence times. Experimental results demonstrate that \texttt{gsplat} achieves up to 10\% less training time and $4\times$ less memory than the original \cite{kerbl20233d} implementation. Utilized in several research projects, \texttt{gsplat} is actively maintained on GitHub. Source code is available at \url{https://github.com/nerfstudio-project/gsplat} under Apache License 2.0. We welcome contributions from the open-source community.
\end{abstract}


\section{Introduction}%
Gaussian Splatting, a seminal work proposed by \cite{kerbl20233d} is a rapidly developing area of research for high fidelity 3D scene reconstruction and novel view synthesis with wide interest in both academia and industry. It outperforms many of the previous NeRF-based \citep{mildenhall2020nerf} methods in several important areas, including i) computational efficiency for training and rendering, ii) ease of editing and post-processing, and iii) deployability on hardware-constrained devices and web-based technologies. In this paper, we introduce \texttt{gsplat}, an open-source project built around Gaussian Splatting that aims to be an efficient and user-friendly library. The underlying concept is to enable a simple and easily modifiable API for PyTorch-based projects developing Gaussian Splatting models. \texttt{gsplat} supports the latest research features and is developed with modern software engineering practices in mind. Since its initial release in October 2023, \texttt{gsplat} has garnered 39 contributors and over 1.6k stars on GitHub. \texttt{gsplat} is released under the Apache License 2.0. Documentation and further information are available on the website at:
\begin{center}
  \url{http://docs.gsplat.studio/}
\end{center}
The closest prior work implementing open-source Gaussian Splatting methods include GauStudio \citep{ye2024gaustudio} which consolidates various research efforts into a single code repository and several PyTorch-based reproductions \citep{januschGaussianCuda, torch-splatting}. Unlike previous work, \texttt{gsplat} not only seeks to implement the original 3DGS work with performance improvements, but aims to provide an easy-to-use and modular API interface allowing for external extensions and modifications, promoting further research in Gaussian Splatting. We welcome contributions from students, researchers, and the open-source community.

\section{Design}%
\texttt{gsplat} is a standalone library developed with efficiency and modularity in mind.  It is installed from PyPI on both Windows and Linux platforms, and provides a PyTorch interface. For speed considerations, many operations are programmed into optimized CUDA kernels and exposed to the developer via Python bindings. In addition, a native PyTorch implementation is also carried in \texttt{gsplat} to support iteration on new research ideas. \texttt{gsplat} is designed to provide a simple interface that can be imported from external projects, allowing easy integration of the main Gaussian Splatting functionality as well as algorithmic customization based on the latest research. With well-documented examples, test cases verifying the correctness of CUDA operations, and further documentation hosted online, \texttt{gsplat} can also serve as an education resource for new researchers entering the field.
\hspace{1em}
\begin{minipage}[h!]{0.7\textwidth}
\lstset{style=codeblock}
\begin{lstlisting}[language=Python,]
import torch
from gsplat import rasterization
# Initialize a 3D Gaussian:
mean = torch.tensor([[0.,0.,0.01]], device="cuda")
quat = torch.tensor([[1.,0.,0.,0.]], device="cuda")
color = torch.rand((1, 3), device="cuda")
opac = torch.ones((1,), device="cuda")
scale = torch.rand((1, 3), device="cuda")
view = torch.eye(4, device="cuda")[None]
K = torch.tensor([[[1., 0., 120.], [0., 1., 120.], [0., 0., 1.]]], device="cuda") # camera intrinsics
# Render an image using gsplat:
rgb_image, alpha, metadata = rasterization(
    mean, quat, scale, opac, color, view, K, 240, 240)
\end{lstlisting}
\end{minipage}
\begin{minipage}[h!]{0.3\textwidth}
    \includegraphics[width=\textwidth]{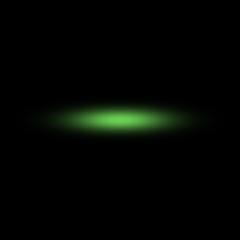}
\end{minipage}
\captionof{figure}{Implementation of the main 3D Gaussian rendering process using the \texttt{gsplat} (v1.3.0) library with only 13 lines of code. A single Gaussian is initialized (left codeblock) and rendered as an RGB image (right).}
\label{fig:1}

\section{Features}
The \texttt{gsplat} librarconsists of features and algorithmic implementations relating to Gaussian Splatting. With a modular interface, users can choose to enable features with simple API calls. Here, wy briefly describe some of the algorithmic enhancements provided by \texttt{gsplat} which are not present in the original 3DGS implementation by \cite{kerbl20233d}.

\boldparagraph{Densification strategies.} A key component of the Gaussian Splatting optimization procedure consists of densification and pruning of Gaussians in under- and over-reconstructed regions of the scene respectively. This has been an active area of research, and the \texttt{gsplat} library supports some of the latest densification strategies. These include the Adaptive Density Control (ADC) proposed by \cite{kerbl20233d}, the Absgrad method proposed in \cite{ye2024absgs}, and the Markov Chain Monte Carlo (MCMC) method proposed in \cite{kheradmand20243d}. \texttt{gsplat}'s modular API allows users to easily change between strategies. For further details regarding densification strategies, we refer to \ref{appendix:densification-strategies}.
\begin{center}
    \begin{minipage}[h!]{0.8\textwidth}%
    \lstset{style=codeblock}
    \begin{lstlisting}[language=Python]
    from gsplat import MCMCStrategy, rasterization
    strategy = MCMCStrategy() #Initialize the strategy 
    strategy_state = strategy.initialize_state()
    for step in range(1000): # Training loop
        render_image, render_alpha, info = rasterization(...)
        strategy.step_pre_backward(...)# Pre-backward step
        loss = ... # Compute the loss
        loss.backward() # Backward pass
        strategy.step_post_backward(...) # Post-backward step
    \end{lstlisting}
    \end{minipage}
    \caption{Code-block for training a Gaussian model with a chosen densification strategy.}
\end{center}

\boldparagraph{Pose optimization.}
\label{Pose optimization}
The Gaussian rendering process (seen in \autoref{fig:1}) in \texttt{gsplat} is fully differentiable, enabling gradient flow to Gaussian parameters $\mathcal{G}(c, \Sigma, \mu, o)$ and to other parameters such as the camera view matrices $\mathcal{P} = [\boldsymbol{R} \mid \boldsymbol{t}]$, which were not considered in the original work. This is crucial for mitigating pose uncertainty in datasets. Specifically, gradients of the reconstruction loss are computed with respect to the rotation and translation components of the camera view matrix, allowing for optimization of initial camera poses via gradient descent. More details are in \ref{appendix:pose-optimization}.

\boldparagraph{Depth rendering.}
\label{Depth rendering}
Rendering depth maps from a Gaussian scene is important for applications such as regularization and meshing. \texttt{gsplat} supports rendering depth maps using an optimized RGB+Depth rasterizer that is also fully differentiable. \texttt{gsplat} supports rendering depth maps using the accumulated z-depth for each pixel and the alpha normalized expected depth. Definitions are found in \ref{appendix:depth-rendering}.

\boldparagraph{N-Dimensional rasterization.}
\label{n-dimensional}
In addition to rendering three-channel RGB images, \texttt{gsplat} also supports rendering higher-dimensional feature vectors. This is motivated by algorithms that combine learned feature maps with differentiable volume rendering \citep{kobayashi2022distilledfeaturefields, lerf2023}. To accommodate the storage needs of these features, the \texttt{gsplat} backend allows for adjustments to parameters affecting memory allocation during training, such as kernel block sizes.

\boldparagraph{Anti-aliasing.}
\label{aliasing}
Viewing a 3D scene represented by Gaussians at varying resolutions can cause aliasing effects, as seen in prior 3D representations \citep{barron2021mipnerf, barron2022mipnerf360}. When the resolution decreases or the scene is viewed from afar, individual Gaussians smaller than a pixel in size produce aliasing artifacts due to sampling below the Nyquist rate. Mip-Splatting \citep{Yu2023MipSplatting} proposes a low pass filter on projected 2D Gaussian covariances, ensuring a Gaussian's extent always spans a pixel.
\texttt{gsplat} supports rendering with the 2D anti-aliasing mode introduced in \citeauthor{Yu2023MipSplatting}. Definitions are found in \ref{appendix:anti-aliasing}

\section{Evaluation}
\label{Evaluation}

\paragraph{Overall comparison.} We compare the training performance and efficiency of \texttt{gsplat} training against the original implementation by \citeauthor{kerbl20233d} on the MipNeRF360 dataset \citep{barron2022mipnerf360}. We use the standard ADC densification strategy and equivalent configuration settings for both. We report average results on novel-view synthesis, memory usage, and training time using an A100 GPU (\texttt{PyTorch} v2.1.2 and \texttt{cudatoolkit} v11.8) at 7k and 30k training iterations in \autoref{table:gsplat_vs_inria}.

\input{tables/mipnerf/gsplat_vs_inria}

We achieve the same rendering performance as the original implementation whilst using less memory and significantly reducing training time. 

\paragraph{Feature comparison.} Furthermore, we analyze the impact of features provided in the \texttt{gsplat} library in \autoref{tab:all_average_metrics}. Additional quantitative evaluations can be found in \autoref{appendix:additional evaluations}.
\begin{table}[h]
    \vspace{-1em}
    \centering
    \caption{\texttt{gsplat} feature comparison on the MipNeRF360 dataset averaged over 7 scenes.}
    \label{tab:all_average_metrics}
    \input{tables/mipnerf/all_average_metrics}
    \vspace{-1em}
\end{table}
\vspace{-1em}
\acks{We thank the many open-source users for their valuable contributions to \texttt{gsplat}: fwilliams (Francis Williams), niujinshuchong (Zehao Yu), and FantasticOven2 (Weijia Zeng). This project was funded in part by NSF:CNS-2235013 and IARPA DOI/IBC No. 140D0423C0035; MT was funded by the Finnish Center for Artificial Intelligence (FCAI); JK and BY are supported by the NSF Research Fellowship Program, Grant DGE 2146752.}


\vskip 0.2in
\bibliography{gsplat}

\newpage

\appendix
\input{supplement}

\end{document}

%% file: tables/mipnerf/gsplat_vs_inria.tex
\begin{table}[h]
\centering
\caption{Comparison of \texttt{gsplat} training performance with the original 3DGS (\citeauthor{kerbl20233d}) implementation on the MipNeRF360 dataset. Results are averaged over 7 scenes.}
\begin{tabular}{l|c|c|c|c|c}
\toprule
 & PSNR $\uparrow$ & SSIM $\uparrow$ & LPIPS $\downarrow$ & Memory $\downarrow$ &  Time (min) $\downarrow$\\
\midrule
\textsc{3DGS -7k} & 27.23 & 0.83 & 0.20 & 7.7 GB & 4.64 \\
\textsc{gsplat -7k} & 27.23 & 0.83 & 0.20 & 4.3 GB & 3.36 \\
\textsc{3DGS -30k} & 28.95 & 0.87 & 0.14 & 9.0 GB & 26.19 \\
\textsc{gsplat -30k} & 29.00 & 0.87 & 0.14 & 5.6 GB & 19.39 \\
\bottomrule
\end{tabular}
\label{table:gsplat_vs_inria}
\end{table}

%% file: tables/mipnerf/all_average_metrics.tex
%
%
\begin{tabular}{l|c|c|c|c|c|c}
\toprule & PSNR$\uparrow$ & SSIM$\uparrow$ & LPIPS$\downarrow$ & Num GS & Mem $\downarrow$ & Time (min) $\downarrow$\\
\midrule
\textsc{gsplat} & 29.00 & 0.87 & 0.14 & 3.24 M & 5.62 GB & 19.39\\
\midrule
 w/ \textsc{absgrad} & 29.11 & 0.88 & 0.12 & 2.47 M & 4.40 GB & 18.10\\
 w/ \textsc{mcmc} & 29.18 & 0.87 & 0.14 & 1.00 M & 1.98 GB & 15.42\\
 w/ \textsc{antialiased} & 29.03 & 0.87 & 0.14 & 3.38 M & 5.87 GB & 19.52\\
\bottomrule
\end{tabular}

%% file: supplement.tex
\section*{Supplementary Material}
In this supplementary material we provide further details regarding the features present in the \texttt{gsplat} library in \autoref{appendix:gsplat-features}. We give additional quantitative comparisons in \autoref{appendix:additional evaluations}. Furthermore, we present additional details regarding the mathematical implementation of the forward pass in \autoref{appendix:forward pass} and backward pass in \autoref{appendix:backward pass}, which are at the core of the \texttt{gsplat} library. Lastly, we explain conventions used in the \texttt{gsplat} library in \autoref{appendix:data conventions}.

\texttt{gsplat} is constantly being updated and improved. For example, recent enhancements have enabled multi-GPU training support for large-scale scene reconstruction. For most recent updates, check the commit history at \url{https://github.com/nerfstudio-project/gsplat}.

\section{Further Details for \texttt{gsplat} Features}
\label{appendix:gsplat-features}
\subsection{Densification Strategies}
\label{appendix:densification-strategies}
As of July 2024, \texttt{gsplat} supports the following densification strategies.

\subsubsection{ADC}
The Adaptive Density Control (ADC) method was originally proposed by \cite{kerbl20233d}. During training, the positional gradients $\nabla_{\tilde{\mu}_n} \mathcal{L}=\|\frac{\partial \mathcal{L}}{\partial \boldsymbol{\tilde{\mu}_n}}\|$ are tracked for a single Gaussian primitive $\mathcal{G}_n(\boldsymbol{\mu}_n, \boldsymbol{\Sigma}_n, \boldsymbol{}{c}_n, o_n)$ and average over multiple renderings with camera views $\{\mathcal{P}\}_{k=1}^M$. If the accumulated positional gradients for a primitive exceed a user set threshold $\mathcal{T}$ (default is $0.0002$), a Gaussian is either split or cloned. Gaussians are split if the extent of the primitive, measured by the size of the largest scale of a Gaussian, is beyond another threshold (set to $0.01$); otherwise, the Gaussian is simply cloned.

\indent The ADC system periodically culls Gaussian primitives based on their opacity values, $o_n$. Gaussians with opacity values below a threshold (set at $0.005$) are removed at fixed intervals during training. Additionally, the ADC system periodically resets all Gaussian opacities to a small value to further encourage the culling of more Gaussians during training.

\subsubsection{Absgrad}
\label{appendix:absgrad}
In the ADC densification strategy, the view space positional gradients for a Gaussian $\nabla_{\tilde{\mu}_n} \mathcal{L} =  \sum_{k=1}^{M} ( \frac{\delta \mathcal{L}_x}{\delta \tilde \mu_x}, \frac{\delta \mathcal{L}_y}{\delta \tilde \mu_y} )$ are tracked across $M$ camera views during training and a criterion for splitting and duplicating is set by a threshold. \citeauthor{ye2024absgs} and \citeauthor{liu2024efficientgs} discovered that this formulation of positional gradient accumulation can result in gradient collisions, where negative and positive view-space gradients cancel each other out, resulting in a poor densification heuristic. They propose to accumulate gradients using absolute sums $\nabla_{\tilde{\mu}_n} \mathcal{L} =  \sum_{k=1}^{M} ( \vert \frac{\delta \mathcal{L}_x}{\delta \tilde \mu_x} \vert, \vert \frac{\delta \mathcal{L}_y}{\delta \tilde \mu_y} \vert)$ instead. \texttt{gsplat} supports training with both versions of view-space accumulated gradients. The Absgrad feature is enabled with a simple API call:
\begin{center}
    \begin{minipage}[h!]{0.8\textwidth}%
    \lstset{style=codeblock}
    \begin{lstlisting}[language=Python]
    for step in range(1000): # Training loop
        rgb_image, alpha, meta_data = rasterization(
            ..., 
            absgrad = True) # Absgrad feature is enabled
        loss = ... 
        loss.backward()
    \end{lstlisting}
    \end{minipage}
    \caption{Training with the Absgrad view space gradients enabled.}
\end{center}

\subsubsection{MCMC}
\label{mcmc}
The authors in \cite{kheradmand20243d} adopt an alternative Bayesian perspective to the densification strategy in Gaussian Splatting. The authors reformulate Gaussian Splatting densification as a Stochastic Gradient Langevin Dynamic (SGLD) update rule and rewrite stochastic gradient descent updates, expressed as with    
$
\mathcal{G} \leftarrow \mathcal{G}-\lambda_{\operatorname{lr}} \cdot \nabla_{\mathcal{G}} \mathbb{E}_{\mathbf{I} \sim \mathcal{I}}\left[\mathcal{L}(\mathcal{G} ; \mathbf{I})\right]
$
as SGLD updates
\begin{equation}
\mathcal{G} \leftarrow \mathcal{G} -\lambda_{\text{lr}} \cdot \nabla_{\mathcal{G}} \mathbb{E}_{\mathbf{I} \sim \mathcal{I}}\left[\mathcal{L}_{\text {total }}(\mathcal{G} ; \mathbf{I})\right]+\lambda_{\text {noise}} \cdot \boldsymbol{\epsilon}
\end{equation}

controlled by hyperparameters $\lambda_{\text {noise}}$ and $\lambda_{\text {lr}}$ and a noise term $\boldsymbol{\epsilon}$ applied to the center locations $\mu$ of Gaussians. 

\subsection{Pose optimization}
\label{appendix:pose-optimization}
Gradients of the reconstruction loss are computed to the rotation and translation components of a given camera view matrix using:
\begin{equation}
    \frac{\delta \mathcal{L}}{\delta \boldsymbol{t}} = -\sum_n \frac{\delta \mathcal{L}}{\delta \boldsymbol{\tilde \mu}_n}, \quad
    \frac{\delta \mathcal{L}}{\delta \boldsymbol{R}} = -\left[\sum_n  \frac{\delta \mathcal{L}}{\delta \boldsymbol{\tilde \mu}_n} (\boldsymbol{\mu}_n -\boldsymbol{t})^\intercal ]\right] \boldsymbol{R}
\end{equation}

\subsection{Depth rendering}
The definitions for accumulated depth and expected depth at a pixel $(x,y)$ are
\label{appendix:depth-rendering}
\begin{center}
    \begin{minipage}{0.1\textwidth}
        \texttt{Accumulated depth}
    \end{minipage}
    \begin{minipage}{0.385\textwidth}
      \begin{equation}
          {d}_{x,y}^{acc} = \sum_{n=1}^{N}  z_n \cdot \alpha_n \cdot T_n
      \end{equation}
    \end{minipage}
    \begin{minipage}{0.1\textwidth}
        \texttt{Expected depth}
    \end{minipage}
    \begin{minipage}{0.385\textwidth}
         \begin{equation}
          {d}_{x,y}^{exp} = \frac{\sum_{n=1}^{N}  z_n \cdot \alpha_n \cdot T_n}{\sum_{n=1}^{N} \alpha_n \cdot T_n}
      \end{equation}
    \end{minipage}
\end{center}
where $T_n = \prod_{j = 1}^{n - 1} (1- \alpha_j) $ is the accumulated transparency of depth-sorted Gaussians at pixel $(x,y)$. 

\subsection{Anti-aliasing}
\label{appendix:anti-aliasing}
\texttt{gsplat} supports rendering under the classic and anti-alias modes which modify the screen-space 2D gaussian sizes $\mathcal{G}^{2D}$ as follows:
\begin{center}
    \begin{minipage}{0.1\textwidth}
        \texttt{Classic mode}
    \end{minipage}
    \begin{minipage}{0.85\textwidth}
      \begin{equation}
          \mathcal{G}^{2D} = o_n \cdot \exp{\left(-\frac{1}{2}(\boldsymbol{p}-\boldsymbol{\mu}_n)^{\intercal} (\boldsymbol{\Sigma}_{n}^{2D} + s\cdot \mathbf{I})^{-1}(\boldsymbol{p}-\boldsymbol{\mu}_n)\right)}
      \end{equation}
    \end{minipage}
    \begin{minipage}{0.1\textwidth}
         \texttt{Anti-alias mode}
    \end{minipage}
    \begin{minipage}{0.85\textwidth}
      \begin{equation}
          \mathcal{G}^{2D} = \sqrt{\frac{\vert \boldsymbol{\Sigma}_{n}^{2D} \vert}{\vert \boldsymbol{\Sigma}_{n}^{2D} + s\cdot \mathbf{I}\vert}} \cdot o_n \cdot \exp{\left(-\frac{1}{2}(\boldsymbol{p}-\boldsymbol{\mu}_n)^{\intercal} (\boldsymbol{\Sigma}_{n}^{2D} + s\cdot \mathbf{I})^{-1}(\boldsymbol{p}-\boldsymbol{\mu}_n)\right)}
      \end{equation}
    \end{minipage}
\end{center}
where $s$ is set as a hyper-parameter during training, default is 0.3, to ensure that a 2D Gaussian's size spans the width of a single pixel.
\section{Additional Evaluations}
\label{appendix:additional evaluations}
We provide additional quantitative evaluation for the various features provided in the \texttt{gsplat} library. We ablate performance using default settings, with Absgrad and MCMC densification strategies, as well as using antialiased rendering. We report per scene novel-view synthesis metrics on the MipNeRF360 dataset in \autoref{table:psnr_mipnerf360}, \autoref{table:ssim_mipnerf360}, and \autoref{table:lpips_mipnerf360} as well as memory usage in \autoref{table:mem_mipnerf360}.

\begin{table}[h!]
    \centering
    \caption{Per scene PSNR metrics on the MipNeRF360 dataset.}
    \input{tables/mipnerf/psnr_per_scene_metrics}
    \label{table:psnr_mipnerf360}
\end{table}
\begin{table}[h!]
    \centering
    \caption{Per scene SSIM metrics on the MiPNeRF360 dataset.}
    \input{tables/mipnerf/ssim_per_scene_metrics}
    \label{table:ssim_mipnerf360}
\end{table}
\begin{table}[h!]
    \centering
    \caption{Per scene LPIPS metrics on the MipNeRF360 dataset. LPIPS is computed using AlexNet features.}
    \input{tables/mipnerf/lpips_per_scene_metrics}
    \label{table:lpips_mipnerf360}
\end{table}
\begin{table}[h!]
    \centering
    \caption{Per scene memory consumption (in GB) metrics on the MipNeRF360 dataset.}
    \input{tables/mipnerf/mem_per_scene_metrics}
    \label{table:mem_mipnerf360}
\end{table}

\input{figures/gradient_flow_diagram}
\section{Forward Pass}
\label{appendix:forward pass}
A 3D Gaussian is parametrized by its mean $\boldsymbol{\mu} \in \mathbb{R}^3$, covariance matrix $\boldsymbol{\Sigma} \in \mathbb{R}^{3\times3}$ decomposed into a scaling vector $\boldsymbol{s} \in \mathbb{R}^{3}$ and a rotation quaternion $\boldsymbol{q} \in \mathbb{R}^4$, opacity $o \in \mathbb{R}$, and a feature vector $\boldsymbol{c} \in \mathbb{R}^{N}$. For the remainder of the derivations, we denote $\boldsymbol{c} \in \mathbb{R}^{3}$ as color encoded via spherical harmonics similar to the original work by \citet{kerbl20233d}; however, the derivations also apply to other $N$-dimensional vectors. 
To render a view from the Gaussian scene, we compute their projected 2D means and extents in the camera
plane. Visible 2D Gaussians are then sorted by depth and composited from front to back using the discrete rendering equation to construct the output image.

\subsection{Projection of Gaussians}
The render camera is described by its extrinsics $\mathcal{P}$, which transforms points from the world coordinate space to the camera coordinate space, and its intrinsics $\mathcal{K}$ which projects Gaussians from camera coordinates to image coordinates:
\begin{equation}
\label{eq: pose and intrinsics}
    \mathcal{P} =
    \left[
    \begin{matrix}
        R & \;t \\
        0 & \;1 \\
    \end{matrix}
    \right],
    \quad
    \mathcal{K} =
    \begin{bmatrix}
        fy & 0 & cx\\
        0 & fy & cx\\
        0 & 0 & 1 \\
    \end{bmatrix}
\end{equation}

\noindent A visible 3D Gaussians $\mathcal{G}_{n}(\boldsymbol{\mu}, \boldsymbol{\Sigma}, o, \boldsymbol{c})$ in world space is mapped into camera space using:
\begin{equation}
\label{eq:project-gaussians-equations}
  \boldsymbol{\hat \mu_{n}} = R^\top (\mu_n - p), \quad
  \boldsymbol{\hat \Sigma_{n}} = R^\top \Sigma R, \quad
  \boldsymbol{\hat c_{n}} = \texttt{SH}(\frac{\mu_n - t}{\Vert{\mu_n - t\Vert}})
\end{equation}

\noindent Furthermore, the camera coordinate Gaussian $\mathcal{\hat G}_n (\boldsymbol{\hat \mu_n}, \boldsymbol{\hat \Sigma_n}, o_n, \boldsymbol{\hat c_n})$ projects to a image space 2D Gaussian $\mathcal{\hat G}_n^{2D}(\boldsymbol{\mu'}, d, \boldsymbol{\Sigma'})$ with z-depth $d$ via:
\begin{equation}
  d = \tilde \mu_z, \quad
  \mu' = (\tilde \mu_x / d, \tilde \mu_y / d), \quad
  \Sigma' = J^\top \hat \Sigma J
\end{equation}

\noindent We approximate the projection of camera space $\boldsymbol{\hat \Sigma_{n}}$ to image space with a first-order Taylor expansion located at the pose $\mathcal{P}$. Specifically, we compute the affine transform $J\in \mathbb{R}^{2\times 3}$ as:
\begin{equation}
\label{eq:pinhole-projection-jacobian}
  J = \frac1{d}
  \left[
    \begin{matrix}
      1 & \;0 &\; -\tilde \mu_x/d \\
      0 & \;1 &\; -\tilde \mu_y/d \\
    \end{matrix}
  \right]
\end{equation}

\noindent Note,  unlike the original implementation by \cite{kerbl20233d}, we do not use the OpenGL NDC coordinate system as an intermediate representation. Thus, a 2D Gaussian \newline $\mathcal{G}_n^{2D} (\boldsymbol{\mu'}, \boldsymbol{\Sigma'}, o, \boldsymbol{c})$ is defined in image coordinates with the covariance matrix $\boldsymbol{\Sigma}' \in \mathbb{R}^{2\times2}$: 
\begin{equation}
    \boldsymbol{\Sigma}_{n}^{2D} = J^\top R^\top \boldsymbol{\Sigma} R J.
\end{equation}
We further map from image to pixel coordinates for rasterization. See \autoref{appendix:data conventions} for more details. 

\subsection{Rasterization of Gaussians}
We directly follow the tile sorting method introduced by \citeauthor{kerbl20233d}, which bins the 2D Gaussians into $16\times16$ tiles and sorts them per tile by depth. For each Gaussian, we compute the axis-aligned bounding box around the 99th percentile ellipse of each 2D projected covariance (3 standard deviations), and include it in a tile
bin if its bounding box intersects with the tile. We then apply the tile sorting algorithm as presented in Appendix C of \cite{kerbl20233d} to get a list of Gaussians sorted by depth for each tile. We then rasterize the sorted Gaussians within each tile. For a color at a pixel $\boldsymbol{p}_{(x,y)}$, let $i$ index the $N$ Gaussians involved in that pixel.
\begin{align}
  {\bf \hat{C}_{x,y}}  & = \sum_{n \in N} {\bf{c}}_{n}\alpha_n T_i, \quad \textrm{ where } T_i = \prod_{j = 1}^{i - 1} (1- \alpha_j)
\end{align}

\noindent We compute $\alpha_n$ with the 2D covariance $\boldsymbol{\Sigma}_{n}^{2D} \in \mathbb{R}^{2\times 2}$ and opacity parameters:
\begin{align}
\label{gaussian-density}
  {\alpha_n}  & = o_n \cdot \exp{\left(-\frac{1}{2}(\boldsymbol{p}_{(x,y)}-\boldsymbol{\mu}_n)^\intercal (\boldsymbol{\Sigma}_n^{2D})^{-1}(\boldsymbol{p}_{(x,y)}-\boldsymbol{\mu}_n)\right)}
\end{align}

\noindent We compute $T_i$ online as we iterate through the Gaussians front to back.

\section{Backward Pass}
\label{appendix:backward pass}
\subsection{Computing Gradients of Gaussians}
We now compute the gradients of a scalar loss with respect to the input Gaussian parameters. That is, given the gradient of a scalar loss $\mathcal{L}$ with respect to each pixel of the output image, we propagate the gradients
backward toward the original input parameters with standard chain rule mechanics. In the following we will use the Frobenius inner product in deriving the matrix derivatives \cite{Giles2008AnEC}:
\begin{equation}
    \langle X, Y \rangle =\textrm{Tr}(X^\top Y) = \textrm{vec}(X)^\top \textrm{vec}(Y) \in \mathbb{R}
\end{equation}

\noindent and can be thought of as a matrix dot product. The Frobenius inner product has the following properties:
\begin{equation}
    \langle X, Y \rangle = \langle Y, X \rangle
\end{equation}
\begin{equation}
    \langle X, Y \rangle = \langle X^\top, Y^\top \rangle,
\end{equation}
\begin{equation}
    \langle X, YZ \rangle = \langle Y\top X, Z \rangle = \langle X Z^\top, Y \rangle,
\end{equation}
\begin{equation}
    \langle X, Y + Z \rangle = \langle X, Y \rangle  + \langle X, Z \rangle
\end{equation}

\noindent Suppose we have a scalar function $f$ of $X \in \mathbb{R}^{m\times n}$, and that $X = AY
$, with $A \in \mathbb{R}^{m\times p}$ and $Y \in \mathbb{R}^{p\times n}$. We can write the gradient of $f$ with respect to an arbitrary scalar $x \in \mathbb{R}$ as
\begin{equation}
    \frac{\partial f}{ \partial x } = \langle \frac{\partial f}{\partial X}, \frac{\partial X}{\partial x}\rangle,
\end{equation}
\noindent for which we use the shorthand
\begin{equation}
    \partial f = \langle \frac{\partial f}{\partial X}, \partial X \rangle.
\end{equation}

\noindent Here $\frac{\partial f}{\partial x} \in \mathbb{R}$, $\frac{\partial f}{\partial X} \in \mathbb{R}^{ m \times n}$, and $\frac{\partial X}{\partial x} \in \mathbb{R}^{ m \times n}$.

\noindent In this case, it is simple to continue the chain rule. Letting $G = \frac{\partial f}{\partial X}$ , we have
\begin{equation*}
    \begin{split}
        \frac{\partial f}{\partial x} & =\left\langle G, \frac{\partial(A Y)}{\partial x}\right\rangle \\
        & =\left\langle G, \frac{\partial A}{\partial x} Y\right\rangle+\left\langle G, A \frac{\partial Y}{\partial x}\right\rangle \\
        & =\left\langle G Y^{\top}, \frac{\partial A}{\partial x}\right\rangle+\left\langle A^{\top} G, \frac{\partial Y}{\partial x}\right\rangle. 
    \end{split}
\end{equation*}

\noindent From here, we read out the elements of the gradients of f with respect to $A$ and $Y$ by letting $x = A_{ij}$ and
$x = Y_{ij}$ respectively, and find that
\begin{equation}
\frac{\partial f}{\partial A}=G Y^{\top} \in \mathbb{R}^{m \times p}, \quad \frac{\partial f}{\partial Y}=A^{\top} G \in \mathbb{R}^{p \times n}
\end{equation}

\subsection{Depth Compositing Gradients}
We start with propagating the loss gradients of a pixel $i$ back to the Gaussians that contributed to the pixel. Specifically, for a Gaussian $n$ that contributes to the pixel $i$, we compute the gradients with respect to color $\frac{\partial \mathcal{L}}{\partial c_n} \in \mathbb{R}^3$, opacity $\frac{\partial \mathcal{L}}{\partial o_n} \in \mathbb{R}$, the $2 \mathrm{D}$ means $\frac{\partial \mathcal{L}}{\partial \mu_n^{\prime}} \in \mathbb{R}^2$, and $2 \mathrm{D}$ covariances $\frac{\partial \mathcal{L}}{\partial \boldsymbol{\Sigma}_n^{\prime}} \in \mathbb{R}^{2 \times 2}$, given the $\frac{\partial \mathcal{L}}{\partial C_i} \in \mathbb{R}^3$. In the forward pass, we compute the contribution of each Gaussian to the pixel color from front to back, i.e. Gaussians in the back are downstream of those in the front. As such, in the backward pass, we compute the gradients of the Gaussians from back to front.
For the color, we have
\begin{equation}
\frac{\partial C_i(k)}{\partial c_n(k)}=\alpha_n \cdot T_n
\end{equation}

for each channel $k$. We save the final $T_N$ value from the forward pass and compute next $T_{n-1}$ values as we iterate backward:
\begin{equation}
T_{n-1}=\frac{T_n}{1-\alpha_{n-1}}
\end{equation}

\noindent For the $\alpha$ gradient, for each channel $k$ we have the scalar gradients
\begin{equation}
\frac{\partial C_i(k)}{\partial \alpha_n}=c_n(k) \cdot T_n-\frac{S_n(k)}{1-\alpha_n} \text { where } S_n=\sum_{m>n} c_m \alpha_m T_m .
\end{equation}

\noindent We can also compute $S_{n-1}$ as we iterate backward over Gaussians:
\begin{equation}
\begin{split}
    S_N(k) & =0 \\
    S_{n-1}(k) & =c_n(k) \alpha_n T_n+S_n(k) .
\end{split}
\end{equation}

\noindent For the opacity and sigma, we have scalar gradients
\begin{equation}
    \frac{\partial \alpha_n}{\partial o_n} = \exp \left(-\sigma_n\right), \quad \frac{\partial \alpha_n}{\partial \sigma_n}=-o_n \exp \left(-\sigma_n\right)
\end{equation}

\noindent For the 2D mean, we have the Jacobian
\begin{equation}
    \frac{\partial \sigma_n}{\partial \mu_n^{\prime}}=\frac{\partial \sigma_n}{\partial \Delta_n}=\Sigma_n^{\prime-1} \Delta_n \in \mathbb{R}^2
\end{equation}

\noindent For the $2 \mathrm{D}$ covariance, we let $Y=\Sigma_n^{\prime-1}$, which has a straightforward Jacobian from $\sigma_n$ :
\begin{equation}
\frac{\partial \sigma_n}{\partial Y}=\frac{1}{2} \Delta_n \Delta_n^{\top} \in \mathbb{R}^{2 \times 2} \text {. }
\end{equation}

\noindent To continue back-propagating through $Y \in \mathbb{R}^{2 \times 2}$, we let $G=\frac{\partial \sigma_n}{\partial Y}$ and write the gradients with respect to a scalar variable $x$ as
\begin{equation}
\frac{\partial \sigma_n}{\partial x}=\left\langle G, \frac{\partial Y}{\partial x}\right\rangle .
\end{equation}

\noindent We use the identity [\cite{petersen}, \cite{dwyer}] that $\frac{\partial Y}{\partial x}=-Y \frac{\partial \Sigma_n^{\prime}}{\partial x} Y$, and have
\begin{equation}
\begin{split}
    \frac{\partial \sigma_n}{\partial x} & =\left\langle G,-Y \frac{\partial \Sigma_n^{\prime}}{\partial x} Y\right\rangle \\
    & =\left\langle-Y^{\top} G Y^{\top}, \frac{\partial \Sigma_n^{\prime}}{\partial x}\right\rangle
\end{split}
\end{equation}

\noindent The gradient of $\sigma_n$ with respect to $\Sigma_n^{\prime}$ is then
\begin{equation}
\frac{\partial \sigma_n}{\partial \Sigma_n^{\prime}}=-\frac{1}{2} \Sigma_n^{\prime-1} \Delta_n \Delta_n^{\top} \Sigma_n^{\prime-1}
\end{equation}

\subsection{Projection Gradients}

Given the gradients of $\mathcal{L}$ with respect the projected $2 \mathrm{D}$ mean $\mu^{\prime}$ and covariance $\Sigma^{\prime}$ of a Gaussian, we can continue to backpropagate the gradients of its $3 \mathrm{D}$ means $\mu$ and covariances $\Sigma$. Here we deal only with a single Gaussian at a time, so we drop the subscript $n$ and compute the gradients $\frac{\partial \mathcal{L}}{\partial \mu} \in \mathbb{R}^3$ and $\frac{\partial \mathcal{L}}{\partial \Sigma} \in \mathbb{R}^{3 \times 3}$, given the gradients $\frac{\partial \mathcal{L}}{\partial \mu^{\prime}} \in \mathbb{R}^2$ and $\frac{\partial \mathcal{L}}{\partial \Sigma^{\prime}} \in \mathbb{R}^{2 \times 2}$.

\noindent We first compute the gradient contribution of $2 \mathrm{D}$ mean $\mu^{\prime}$ to camera coordinates $t \in \mathbb{R}^4$, and of $2 \mathrm{D}$ covariance $\Sigma^{\prime}$ to $3 \mathrm{D}$ covariance $\Sigma$ and camera coordinates $t$. Note that both $\mu^{\prime}$ and $\Sigma^{\prime}$ contribute to the gradient with respect to $t$ :
\begin{equation}
\frac{\partial \mathcal{L}}{\partial t_i}=\frac{\partial \mathcal{L}_{\mu^{\prime}}}{\partial t_i}+\frac{\partial \mathcal{L}_{\Sigma^{\prime}}}{\partial t_i}=\frac{\partial \mathcal{L}}{\partial \mu^{\prime}} \frac{\partial \mu^{\prime}}{\partial t_i}+\left\langle\frac{\partial \mathcal{L}}{\partial \Sigma^{\prime}}, \frac{\partial \Sigma^{\prime}}{\partial t_i}\right\rangle
\end{equation}

\noindent For $2 \mathrm{D}$ mean $\mu^{\prime}$, we have the contribution to the gradient of $t$ as:
\begin{equation}
\frac{\partial \mathcal{L}_{\mu^{\prime}}}{\partial t}=\frac{1}{2} P^{\top}\left[\begin{matrix}
w / t_w & 0 & 0 & -w \cdot t_x / t_w^2 \\
0 & h / t_w & 0 & -w \cdot t_y / t_w^2
\end{matrix}\right]^{\top} \frac{\partial \mathcal{L}}{\partial \mu^{\prime}} .
\end{equation}

\noindent The $2 \mathrm{D}$ covariance $\Sigma^{\prime}$ contributes to the gradients of $\Sigma$ and $t$. where $\Sigma^{\prime}=T \Sigma T^{\top}$. The contribution to $\Sigma$ is straightforward. Letting $G=\frac{\partial \mathcal{L}}{\partial \Sigma^{\prime}}$, we have
\begin{equation}
\begin{aligned}
\partial \mathcal{L}_{\Sigma^{\prime}} & =\left\langle G, \partial \Sigma^{\prime}\right\rangle \\
& =\left\langle G,(\partial T) \Sigma T^{\top}+T(\partial \Sigma) T^{\top}+T \Sigma\left(\partial T^{\top}\right)\right\rangle \\
& =\left\langle G T \Sigma^{\top}, \partial T\right\rangle+\left\langle T^{\top} G T, \partial \Sigma\right\rangle+\left\langle G^{\top} T \Sigma, \partial T\right\rangle \\
& =\left\langle G T \Sigma^{\top}+G^{\top} T \Sigma, \partial T\right\rangle+\left\langle T^{\top} G T, \partial \Sigma\right\rangle .
\end{aligned}
\end{equation}

\noindent We read out the gradient with respect to $\Sigma \in \mathbb{R}^{3 \times 3}$ as
\begin{equation}
\frac{\partial \mathcal{L}}{\partial \Sigma}=T^{\top} \frac{\partial \mathcal{L}}{\partial \Sigma^{\prime}} T \text {. }
\end{equation}

\noindent We continue to propagate gradients through $T=J R_{\mathrm{cw}} \in \mathbb{R}^{2 \times 3}$ for $J \in \mathbb{R}^{2 \times 3}$ :
\begin{equation}
\partial \mathcal{L}=\left\langle\frac{\partial \mathcal{L}}{\partial T},(\partial J) R_{\mathrm{cw}}\right\rangle=\left\langle\frac{\partial \mathcal{L}}{\partial T} R_{\mathrm{cw}}^{\top}, \partial J\right\rangle, \quad \text { where } \frac{\partial \mathcal{L}}{\partial T}=\frac{\partial \mathcal{L}}{\partial \Sigma^{\prime}} T \Sigma^{\top}+\frac{\partial \mathcal{L}}{\partial \Sigma^{\prime}} T \Sigma .
\end{equation}

\noindent We continue propagating through $J$ for camera coordinates $t \in \mathbb{R}^4$ for the contribution through $\Sigma^{\prime}$ to the gradients of $t$ :
\begin{align}
\frac{\partial J}{\partial t_x}=\left[\begin{matrix}
0 & 0 & -f_x / t_z^2 \\
0 & 0 & 0
\end{matrix}\right],
\quad  \frac{\partial J}{\partial t_y}=\left[\begin{matrix}
0 & 0 & 0 \\
0 & 0 & -f_y / t_z^2
\end{matrix}\right], 
\\
\quad 
\frac{\partial J}{\partial t_z}=\left[\begin{matrix}
-f_x / t_z^2 & 0 & 2 f_x t_x / t_z^3 \\
0 & -f_y / t_z^2 & 2 f_y t_y / t_z^3
\end{matrix}\right], 
\quad \frac{\partial J}{\partial t_w}=\mathbf{0}^{2 \times 3} .
\end{align}

\noindent We can now sum the two gradients $\frac{\partial \mathcal{L}_{\mu^{\prime}}}{\partial t}$ and $\frac{\partial \mathcal{L}_{\Gamma^{\prime}}}{\partial t}$ into $G=\frac{\partial \mathcal{L}}{\partial t}$, and compute the full gradients with respect to the $3 \mathrm{D}$ mean $\mu$ and the view matrix $T_{\mathrm{cw}}$. We have that $t=T_{\mathrm{cw}} q$, where $q=\left[\begin{matrix}\mu & 1\end{matrix}\right]^{\top}$.
\begin{equation}
\begin{aligned}
\partial \mathcal{L} & =\langle G, \partial t\rangle=\left\langle G, \partial\left(T_{\mathrm{cw}} q\right)\right\rangle \\
& =\left\langle G q^{\top}, \partial T_{\mathrm{cw}}\right\rangle+\left\langle T_{\mathrm{cw}}^{\top} G, \partial q\right\rangle .
\end{aligned}
\end{equation}

\noindent The gradients with respect to $T_{\mathrm{cw}}$ and $\mu$ are then
\begin{equation}
\frac{\partial \mathcal{L}}{\partial T_{\mathrm{cw}}}=\frac{\partial \mathcal{L}}{\partial t} q^{\top} \in \mathbb{R}^{4 \times 4}, \quad \frac{\partial \mathcal{L}}{\partial \mu}=R_{\mathrm{cw}}^{\top}\left[\begin{matrix}
\frac{\partial \mathcal{L}}{\partial t_x} & \frac{\partial \mathcal{L}}{\partial t_y} & \frac{\partial \mathcal{L}}{\partial t_z}
\end{matrix}\right]^{\top} \in \mathbb{R}^3
\end{equation}

\subsection{Scale and rotation gradients}

Now we have $\Sigma=M M^{\top}$ and $\frac{\partial \mathcal{L}}{\partial \Sigma}$. Letting $G=\frac{\partial \mathcal{L}}{\partial \Sigma}$, we have
\begin{equation}
\begin{aligned}
\partial \mathcal{L} & =\langle G, \partial \Sigma\rangle \\
& =\left\langle G,(\partial M) M^{\top}+M\left(\partial M^{\top}\right)\right\rangle \\
& =\left\langle G M+G^{\top} M, \partial M\right\rangle
\end{aligned}
\end{equation}
which gives us
\begin{equation}
\frac{\partial \mathcal{L}}{\partial M}=\frac{\partial \mathcal{L}}{\partial \Sigma} M+\frac{\partial \mathcal{L}}{\partial \Sigma}^{\top} M
\end{equation}

\noindent Now we have $M=R S$, with $G=\frac{\partial \mathcal{L}}{\partial M}$ as
\begin{equation}
\begin{aligned}
\partial \mathcal{L} & =\langle G, \partial M\rangle \\
& =\langle G,(\partial R) S\rangle+\langle G, R(\partial S)\rangle \\
& =\left\langle G S^{\top}, \partial R\right\rangle+\left\langle R^{\top} G, \partial S\right\rangle
\end{aligned}
\end{equation}
which gives us
\begin{equation}
\frac{\partial \mathcal{L}}{\partial R}=\frac{\partial L}{\partial M} S^{\top}, \quad \frac{\partial \mathcal{L}}{\partial S}=R^{\top} \frac{\partial L}{\partial M} .
\end{equation}

\noindent The Jacobians of the rotation matrix $R$ wrt the quaternion parameters $q=(w, x, y, z)$ are
\begin{align}
\frac{\partial R}{\partial w}=2\left[\begin{array}{ccc}0 & -z & y \\ z & 0 & -x \\ -y & x & 0\end{array}\right], \frac{\partial R}{\partial x}=2\left[\begin{array}{ccc}0 & y & z \\ y & -2 x & -w \\ z & w & -2 x\end{array}\right], 
\\ 
\frac{\partial R}{\partial y}=2\left[\begin{array}{ccc}-2 y & x & w \\ x & 0 & z \\ -w & z & -2 y\end{array}\right], \frac{\partial R}{\partial z}=2\left[\begin{array}{ccc}-2 z & -w & x \\ w & -2 z & y \\ x & y & 0\end{array}\right].
\end{align}

\noindent The Jacobians of the scale matrix $S$ with respect to the scale parameters $s=\left(s_x, s_y, s_z\right)$ are
\begin{equation}
\frac{\partial S}{\partial s_j}=\delta_{i j}
\end{equation}
whichs selects the corresponding diagonal element of $\frac{\partial \mathcal{L}}{\partial S}$.


\section{Data Conventions}
\label{appendix:data conventions}
Various conventions are used within the \texttt{gsplat} library. We briefly outline the most important ones.
\subsubsection{Rotation matrix representation}
Similar to the original work by \citeauthor{kerbl20233d}, we represent a Gaussian rotation by a four dimensional quaternion $ q = (w,x,y,z)$ with the Hamilton convention such that the $SO(3)\in \mathbb{R}^{3 \times 3}$ rotation matrix is given by

\begin{equation}
 R = \begin{bmatrix}
        1 - 2 \left( y^2 + z^2 \right) & 2 \left( x y - w z \right) & 2 \left( x z + w y \right) \\
        2 \left( x y + w z \right) & 1 - 2 \left( x^2 + z^2 \right) & 2 \left( y z - w x \right) \\
        2 \left( x z - w y \right) & 2 \left( y z + w x \right) & 1 - 2 \left( x^2 + y^2 \right) \\
        \end{bmatrix}.
\end{equation}

\subsubsection{Pixel Coordinates}
Conversion to discrete pixel coordinates $\boldsymbol{p} = (p_i,p_j) \in \mathbb{Z}^+$ from continuous image coordinates $\boldsymbol{\mu'} = (\mu'_x,\mu'_y) \in \mathbb{R}^+$ assumes that a pixel's center is located at the center of a box of area 1. This gives the following relation between pixel space, image space, and 3D coordinates $\boldsymbol{t} = (t_x, t_y, t_z)$:

\begin{equation}
\begin{aligned}
    p_i + 0.5 = \mu'_x = f_x \cdot t_x / t_z + c_x \\
    p_j + 0.5 = \mu'_y = f_y \cdot t_y / t_z + c_y
\end{aligned}
\end{equation}

\noindent where $(f_x,f_y,c_x,c_y)$ are the pinhole camera intrinsics.

%% file: tables/mipnerf/psnr_per_scene_metrics.tex
%
%
\begin{tabular}{l|c|c|c|c|c|c|c}
\toprule
 & \textbf{Bicycle} & \textbf{Bonsai} & \textbf{Counter} & \textbf{Garden} & \textbf{Kitchen} & \textbf{Room} & \textbf{Stump}\\
 \midrule
 \textsc{gsplat} & 25.29 & 32.21 & 29.01 & 27.39 & 31.37 & 31.23 & 26.51\\
 \midrule
 \textsc{absgrad} & 25.44 & 31.98 & 29.07 & 27.47 & 31.65 & 31.43 & 26.71\\
 \textsc{mcmc 1 mill} & 25.27 & 32.54 & 29.40 & 27.03 & 31.39 & 32.01 & 26.66\\
 \textsc{mcmc 2 mill} & 25.52 & 32.99 & 29.56 & 27.40 & 31.99 & 32.34 & 26.90\\
 \textsc{mcmc 3 mill} & 25.58 & 33.13 & 29.65 & 27.65 & 32.21 & 32.40 & 26.93\\
 \textsc{antialiased} & 25.31 & 32.27 & 29.01 & 27.33 & 31.34 & 31.53 & 26.44\\
\bottomrule
\end{tabular}

%% file: tables/mipnerf/ssim_per_scene_metrics.tex
%
%
\begin{tabular}{l|c|c|c|c|c|c|c}
\toprule
 & \textbf{Bicycle} & \textbf{Bonsai} & \textbf{Counter} & \textbf{Garden} & \textbf{Kitchen} & \textbf{Room} & \textbf{Stump}\\
 \midrule
 \textsc{gsplat} & 0.77 & 0.94 & 0.91 & 0.87 & 0.93 & 0.92 & 0.77\\
 \midrule
 \textsc{absgrad} & 0.78 & 0.94 & 0.91 & 0.87 & 0.93 & 0.92 & 0.78\\
 \textsc{mcmc 1 mill} & 0.77 & 0.95 & 0.92 & 0.85 & 0.93 & 0.93 & 0.78\\
 \textsc{mcmc 2 mill} & 0.78 & 0.95 & 0.92 & 0.87 & 0.93 & 0.93 & 0.79\\
 \textsc{mcmc 3 mill} & 0.79 & 0.95 & 0.92 & 0.87 & 0.94 & 0.93 & 0.79\\
 \textsc{antialiased} & 0.77 & 0.94 & 0.91 & 0.87 & 0.93 & 0.92 & 0.77\\
\bottomrule
\end{tabular}

%% file: tables/mipnerf/lpips_per_scene_metrics.tex
%
%
\begin{tabular}{l|c|c|c|c|c|c|c}
\toprule
 & \textbf{Bicycle} & \textbf{Bonsai} & \textbf{Counter} & \textbf{Garden} & \textbf{Kitchen} & \textbf{Room} & \textbf{Stump}\\
 \midrule
 \textsc{gsplat} & 0.17 & 0.13 & 0.15 & 0.08 & 0.10 & 0.17 & 0.16\\
 \midrule
 \textsc{absgrad} & 0.14 & 0.13 & 0.15 & 0.07 & 0.09 & 0.15 & 0.14\\
 \textsc{mcmc 1 mill} & 0.20 & 0.12 & 0.14 & 0.11 & 0.10 & 0.15 & 0.17\\
 \textsc{mcmc 2 mill} & 0.17 & 0.12 & 0.13 & 0.09 & 0.09 & 0.14 & 0.15\\
 \textsc{mcmc 3 mill} & 0.15 & 0.11 & 0.13 & 0.08 & 0.09 & 0.14 & 0.14\\
 \textsc{antialiased} & 0.18 & 0.13 & 0.16 & 0.08 & 0.10 & 0.17 & 0.16\\
\bottomrule
\end{tabular}

%% file: tables/mipnerf/mem_per_scene_metrics.tex
%
%
\begin{tabular}{l|c|c|c|c|c|c|c}
\toprule
 & \textbf{Bicycle} & \textbf{Bonsai} & \textbf{Counter} & \textbf{Garden} & \textbf{Kitchen} & \textbf{Room} & \textbf{Stump}\\
 \midrule
 \textsc{gsplat} & 10.47 & 2.41 & 2.36 & 9.89 & 3.16 & 2.84 & 8.20\\
 \midrule
 \textsc{absgrad} & 8.75 & 1.91 & 2.02 & 6.36 & 2.84 & 2.75 & 6.15\\
 \textsc{mcmc 1 mill} & 1.84 & 2.06 & 2.16 & 1.81 & 2.05 & 2.14 & 1.82\\
 \textsc{mcmc 2 mill} & 3.21 & 3.51 & 3.57 & 3.18 & 3.51 & 3.84 & 3.17\\
 \textsc{mcmc 3 mill} & 4.75 & 5.11 & 5.59 & 4.54 & 4.97 & 5.38 & 4.59\\
 \textsc{antialiased} & 11.30 & 2.41 & 2.34 & 10.10 & 3.17 & 2.81 & 8.97\\
\bottomrule
\end{tabular}

%% file: figures/gradient_flow_diagram.tex
\newcommand{\bs}[1]{\boldsymbol{#1}}
\newcommand{\loss}{\mathcal{L}}
\newcommand{\pard}[2]{\frac{\partial#1}{\partial#2}}
\newcommand{\norm}[1]{\left\lVert#1\right\rVert}
\newcommand{\qh}{\hat{q}}
\newcommand{\qn}{q}
\newcommand{\cov}{\boldsymbol{\Sigma}}
\newcommand{\covcam}{\boldsymbol{\Sigma}_c}
\newcommand{\covsplat}{\boldsymbol{\Sigma}'}
\newcommand{\covsplati}[1]{\boldsymbol{\Sigma}_{#1}'}
\newcommand{\mycomment}[1]{}
\newcommand{\matdim}[3]{\underset{\scriptscriptstyle #2\times #3}{#1\vphantom(}}
\newcommand{\vecv}{\hat{\bs{v}}}
\newcommand{\vecp}{\bs{p}}
\newcommand{\vecr}{\bs{r}}
\newcommand{\cam}[1]{#1_{c}}
\newcommand{\vecx}{\bs{x}}
\newcommand{\vecX}{\bs{X}}
\newcommand{\vecc}{\bs{c}}
\newcommand{\std}{\hat{\sigma}}
\newcommand{\vecS}{\bs{S}}
\newcommand{\vecR}{\bs{R}}
\newcommand{\vecW}{\bs{W}}
\newcommand{\vecJ}{\bs{J}}
\newcommand{\vect}{\bs{t}}
\newcommand{\kbar}{\bar{\kappa}}
\newcommand{\kvar}{\sigma_{\kbar}^2}
\newcommand{\CUDA}{\texttt{CUDA}~}
\newcommand{\COLMAP}{\texttt{COLMAP}~}
\newcommand{\splatfacto}{\texttt{splatfacto}~}
\newcommand{\grad}[1]{\nabla\loss_{#1}}
\newcommand{\vecq}{\bs{q}}
\newcommand{\vecqh}{\hat{\bs{q}}}
\newcommand{\vecs}{\bs{s}}

\definecolor{red_axis}{rgb}{0.8,0,0}
\definecolor{green_axis}{rgb}{0.2, 0.7, 0.2}
\definecolor{blue_axis}{rgb}{0,0.4,1}

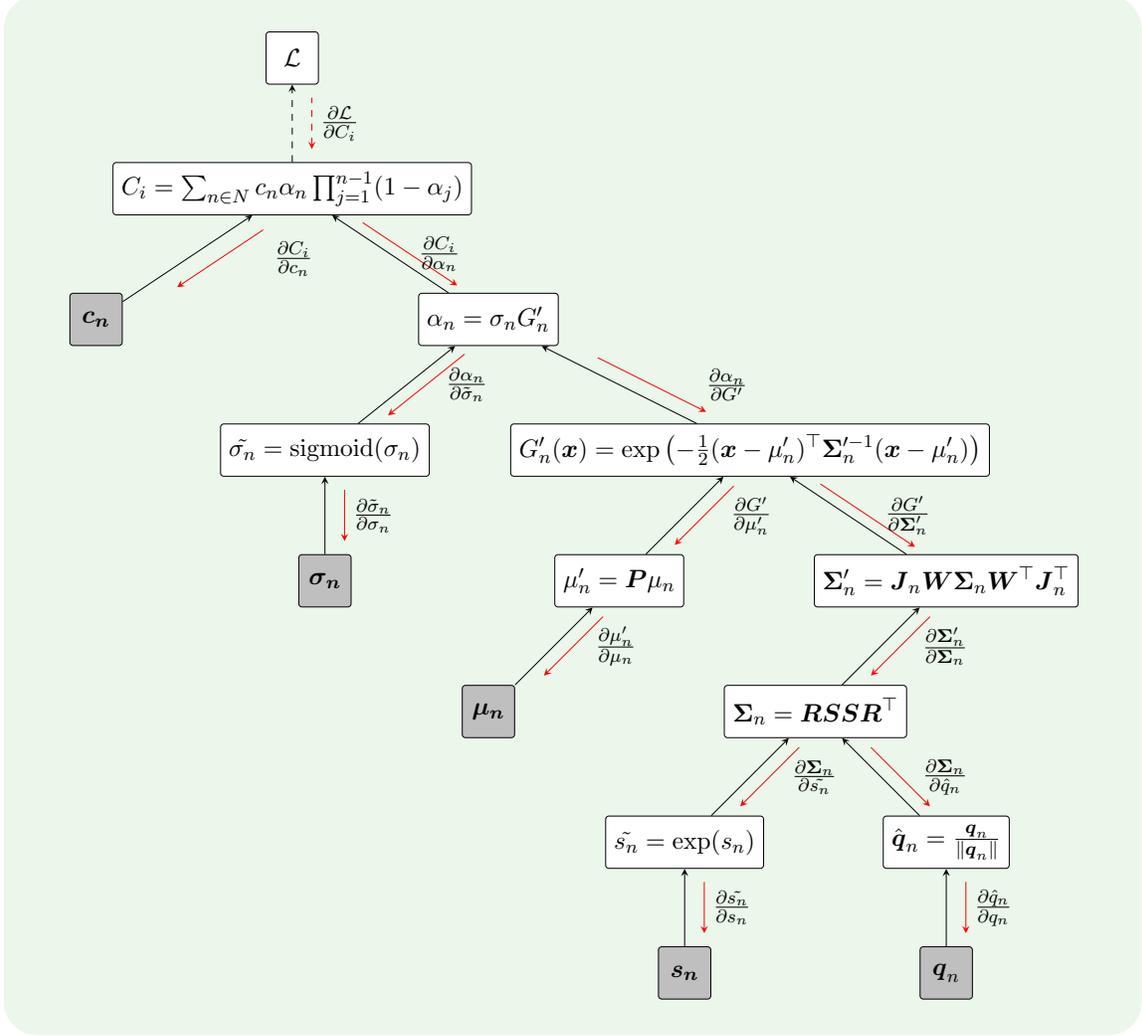
\begin{figure}[h]
    \centering
    \resizebox{1\textwidth}{!}{%
    \begin{tikzpicture} [>=stealth]
        \tikzset{
            inter/.style = {rectangle, rounded corners=0.05cm, minimum height=0.8cm, minimum width=0.8cm, draw = black, fill=white}
        }
        \tikzset{
            leaf/.style = {rectangle, rounded corners=0.05cm, minimum height=0.8cm, minimum width=0.8cm, draw = black, fill=black!25!white}
        }

        \fill[opacity = 0.1, green_axis, rounded corners = 0.5cm] (8.99,0.95) rectangle (-8.41, -14.95);

        \node[inter] (Loss) at (-4,0){$\loss$};
        \node[inter] (C_i) at (-4,-2){$C_i = \sum_{n\in N}c_n\alpha_n\prod_{j=1}^{n-1}(1-\alpha_j)$};
        \node[leaf] (c) at (-7,-4){$\bs{c_n}$};
        \node[inter] (alpha) at (-1,-4){$\alpha_n = \sigma_nG_n'$};
        \node[inter] (tildesigma) at (-3.5,-6){$\tilde{\sigma_n} = \text{sigmoid}(\sigma_n)$};
        \node[leaf] (sigma) at (-3.5,-8){$\bs{\sigma_n}$};
        \node[inter] (gaussian) at (3,-6){$G_n'(\vecx) = \text{exp}\left(-\frac{1}{2}(\vecx-\mu_n')^\top\covsplati{n}^{-1}(\vecx-\mu_n')\right)$};
        \node[inter] (musplat) at (1,-8){$\mu'_n = \bs{P}\mu_n$};
        \node[inter] (covsplat) at (6,-8){$\covsplati{n} = \vecJ_n\vecW\cov_n\vecW^\top\vecJ_n^\top$};
        \node[leaf] (mu) at (-1,-10){$\bs{\mu_n}$};
        \node[inter] (cov) at (4,-10){$\cov_n = \vecR\vecS\vecS\vecR^\top$};
        \node[inter] (tildes) at (2,-12){$\tilde{s_n} = \exp(s_n)$};
        \node[leaf] (s) at (2,-14){$\bs{s_n}$};
        \node[inter] (normquat) at (6,-12){$\hat{\bs{q}}_n = \frac{\bs{q}_n}{\norm{\bs{q}_n}}$};
        \node[leaf] (quat) at (6,-14){$\bs{q}_n$};

        \draw[->] (quat) -- (normquat);
        \draw[->] (s) -- (tildes);
        \draw[->] (tildes) -- (cov);
        \draw[->] (normquat) -- (cov);
        \draw[->] (cov) -- (covsplat);
        \draw[->] (mu) -- (musplat);
        \draw[->] (musplat) -- (gaussian);
        \draw[->] (covsplat) -- (gaussian);
        \draw[->] (gaussian) -- (alpha);
        \draw[->] (sigma) -- (tildesigma);
        \draw[->] (tildesigma) -- (alpha);
        \draw[->] (c) -- (C_i);
        \draw[->] (alpha) -- (C_i);
        \draw[->, dashed] (C_i) -- (Loss);

        \begin{scope}[transform canvas={xshift=0.3cm}]
            \draw[<-, red, shorten <= 0.2cm, shorten >= 0.2cm] (quat) -- (normquat) node[midway, right, black] {$\pard{\hat{q}_n}{q_n}$};
            \draw[<-, red, shorten <= 0.2cm, shorten >= 0.2cm] (s) -- (tildes) node[midway, right, black] {$\pard{\tilde{s_n}}{s_n}$};
            \draw[<-, red, shorten <= 0.2cm, shorten >= 0.2cm] (tildes) -- (cov) node[midway, right, black, outer sep=6pt] {$\pard{\cov_n}{\tilde{s_n}}$};
            \draw[<-, red, shorten <= 0.2cm, shorten >= 0.2cm] (normquat) -- (cov) node[midway, right, black, outer sep=6pt] {$\pard{\cov_n}{\hat{q}_n}$};
            \draw[<-, red, shorten <= 0.2cm, shorten >= 0.2cm] (cov) -- (covsplat) node[midway, right, black, outer sep=6pt] {$\pard{\covsplati{n}}{\cov_n}$};
            \draw[<-, red, shorten <= 0.2cm, shorten >= 0.2cm] (mu) -- (musplat) node[midway, right, black, outer sep=6pt] {$\pard{\mu'_n}{\mu_n}$};
            \draw[<-, red, shorten <= 0.2cm, shorten >= 0.2cm] (musplat) -- (gaussian) node[midway, right, black, outer sep=8pt] {$\pard{G'}{\mu'_n}$};
            \draw[<-, red, shorten <= 0.2cm, shorten >= 0.2cm] (covsplat) -- (gaussian) node[midway, right, black, outer sep=4pt] {$\pard{G'}{\covsplati{n}}$};
            \draw[<-, red, shorten <= 0.2cm, shorten >= 0.2cm] (sigma) -- (tildesigma) node[midway, right, black] {$\pard{\tilde{\sigma}_n}{\sigma_n}$};
            \draw[<-, red, shorten <= 0.2cm, shorten >= 0.2cm] (tildesigma) -- (alpha) node[midway, right, black, outer sep=5pt] {$\pard{\alpha_n}{\tilde{\sigma}_n}$};
            \draw[<-, red, shorten <= 0.2cm, shorten >= 0.2cm] (alpha) -- (C_i) node[midway, right, black] {$\pard{C_i}{\alpha_n}$};
            \draw[<-, red, shorten <= 0.2cm, shorten >= 0.2cm, dashed] (C_i) -- (Loss) node[midway, right, black] {$\pard{\loss}{C_i}$};
        \end{scope}

        \begin{scope}[transform canvas={xshift=0.5cm}]
            \draw[<-, red, shorten <= 0.4cm, shorten >= 0.4cm] (c) -- (C_i) node[midway, right, black, outer sep=20pt] {$\pard{C_i}{c_n}$};
            \draw[<-, red, shorten <= 0.4cm, shorten >= 0.4cm] (gaussian) -- (alpha) node[midway, right, black, outer sep=20pt] {$\pard{\alpha_n}{G'}$};;
        \end{scope}
    \end{tikzpicture}
    }

    \caption{An illustration of the forward (\autoref{appendix:forward pass}) and backward (\autoref{appendix:backward pass}) computation graphs of the main \texttt{gsplat} Gaussian Splatting rendering function for Gaussian parameters $c, \sigma, \mu, s, \text{and }q$.}
    \label{fig:computational_graph}
\end{figure}